# Round Trip Time Prediction Using the Symbolic Function Network Approach

George S. Eskander
*Telecom Egypt*
*Giza, Egypt*
george.samir@telecomegypt.com.eg

Amir Atiya
*Dept Computer Engineering*
*Cairo University*
*Giza, Egypt*
amiratiya@link.net

Kil To Chong
*Chonbuk National University*
*Korea*
kitchong@chonbuk.ac.kr

Hyongsuk Kim
*Chonbuk National University*
*Korea*
hskim@chonbuk.ac.kr

Sung Goo Yoo
*Chonbuk National University*
*Korea*
ding5@chonbuk.ac.kr

## Abstract

*In this paper, we develop a novel approach to model the Internet round trip time using a recently proposed symbolic type neural network model called symbolic function network. The developed predictor is shown to have good generalization performance and simple representation compared to the multilayer perceptron based predictors.*

## 1. Introduction

As being the widest packet switched network, the Internet is targeted to be reliable enough to provide all kinds of services in the best quality, whatever being non real time or real time services. For the non real time applications, the TCP protocol is reliable enough to handle the congestion problems that degrade the transmission performance. On the other hand, for the real time applications, TCP is not suitable due to the play back timing constraints of such kinds of services. Hence, the UDP protocol is being used instead. Although, UDP is suitable for supporting the real time service from the perspective of controlling the transport timing to suit the play back timing needs; it does not consider the quality of service. As a result, a new family of transport algorithms called "TCP-friendly" algorithms has been introduced and still being developed to support real time applications over the Internet [3],[6],[7]. The end–to–end packet delay and the packet loss ratio (PLR) are the key quantities that could be used by such control protocols [5], [8]. The end-to-end packet delay is the delay experienced to deliver a packet over a packet switched network from its original source to the final destination. The delay can be expressed by the round trip time (RTT). However, the packet loss rate is the ratio of the number of lost packets to the total number of sent packets within a specific period. Besides being used as inputs to the transport controllers, these quantities could be used to express and expect the quality of service that may be offered along a specific path between two nodes on the network. These parameters can be measured by using simulators that sends packets from the source to the destination and tracks its delivery. Although designing and running such simulators is an easy task, the accuracy of the measured parameters are not very good due to some noise sources, for example; the difference in synchronization between the source and the destination clocks [28],[29]. Rather than using the measured parameter values directly as inputs to the traffic controllers, it is much better to use the prediction of the future values; as predictive control may give better results than reactive control [4]. To predict these parameters, the system dynamics should be investigated. For the Internet, characterizing the dynamics of such performance parameters is a very challenging issue due to its heterogeneous structure.

In this paper we focus on RTT prediction; however, the statistical analysis shows some correlation between RTT and PLR [26]. So, the prediction of RTT may be used in predicting PLR [27]. There is much work in literature that study the behavior of the end -to -end delay dynamics analytically using the statistical analysis and the queuing theory [10]-[12],[15],[19],[21]-[25]. On the other hand, relatively few empirical methods are applied on this problem that resulted in better performance due to the none stationary nature of the system. Some system identification technique is applied in [18]. Also, some machine learning methods are investigated; a recurrent neural network based predictor is developed in [14] to produce single step and multi step ahead predictions of the RTT time series. A Functional network based RTT predictor is introduced in [30]. A Multiple model approach is developed in [17].

We propose an empirical approach that uses the collected RTT time series to construct a neural type predictor using a new neural network model called symbolic function network (SFN) [1]. This model is proved to have great approximation power and good ability to filter the system noise. Also the constructed symbolic networks are relatively simple and have fast recall speed.

The paper is organized as follows; the next chapter gives an overview on SFN. A synthetic modeling problem is examined that shows the powerful in modeling, noise filtering and simple representation. In Section 3 we design the RTT predictor using the SFN, followed by the conclusion section.

## 2. Overview over the symbolic function network

### 2.1. Theory

The model is designed so as to represent any function or system in a symbolic form in terms of a number of elementary functions. These functions are selected and tuned based on the learning data to form a tree that is built in a constructive way in a top down fashion. Elementary functions are added to the tree one by one in some way so as to achieve as best fitting performance as possible for the given training data. By having several layers of the tree several levels of function concatenations can be achieved. The basic building blocks used are:

$u^v$, $\exp(u)$, and $\log(u)$

That is because most functions encountered in practice involve these basic functions. The constructive algorithm uses these nonlinear transformations as building blocks to synthesize the function to be modeled. Instead of generating a very large number of fixed parameter transformations and let the training process to choose the best smallest set; a small set of flexible transformations are being generated and their parameters are getting tuned by the training process using a suitable optimization technique.

To avoid numerical instabilities and division by zero, we have used instead of the above functions the following modified elementary functions:

$E1(u) = w(u^2+1)^v$, $E2(u) = qe^{\alpha u}$, $E3(u) = p\log(u^2+1)$

Where $u$ is a scalar input. These elements are tunable; where $w, v, q, \alpha,$ and $p$ are the free parameters to be tuned. This scheme offers great modeling flexibility because it permits functional selection, feature selection, and flexible network structure.

### 2.2. Training Algorithms

Several algorithms are proposed to construct the network; that is determining the strategy and sequence of elementary functions additions. In the forward algorithm, the network components are added to the network in an incremental way. The network component can be a single link as in the case of the "Forward Link by Link algorithm" (FLK-SFN), or a complete layer as in the case of the "Forward Layer by Layer algorithm" (FLY-SFN). After adding each element and training the network, the algorithm measures the network performance and decides based on that to keep this added item or to restore the previous network configurations. The number of candidate elements to be added to the network is increasing with the modeled system complexity; so, in case of complex systems the training algorithm will consider a large number of elements that needs long training time. As a result, a model variant called "Forward with Reduced random Set algorithm" (FRS-

SFN) is designed to select a distinct random set of candidate elements to be considered according to some reduction factor determined by the designer. Some model variants are designed that is based on running a pruning algorithm in parallel with the constructive algorithm. By applying pruning, there is a chance to re-evaluate the already admitted elementary functions and removes the redundant links. Also, it can enhance the generalization performance by removing the functions that approximate the system noise, i.e. reducing overfitting. In the Forward-Backward algorithm (FB-SFN), the burning procedure-the backward phase- starts to run after admitting a specific number of elementary functions and then the network construction is being resumed in the forward direction. The forward and backward routines continue alternatively until reaching either the performance goal or the maximum network depth. In the Backward algorithm (B-SFN), the forward phase continues until reaching the maximum network depth and then the backward phase starts.

**2.3. Synthetic Example**

To show the approximation power of the SFN model we investigate here a regression problem, namely approximating a noisy two-dimensional nonlinear function.

**2.3.1. Problem definition.** We considered the function $f(X) = X1^2 * X2^2 + \varepsilon;$ where $\varepsilon$ is a normally distributed random noise with mean $\mu = 0,$ and standard deviation $\sigma = 0.05$.

**2.3.2. Simulation setup.** The training set consists of 100 samples generated in the grid [-1:0.2:0.8]x[-1:0.2:0.8], and the test set consists of 100 samples generated using the grid [-0.975:0.2:0.825]x[-0.975:0.2:0.825]. (I.e. the testing points are shifted from the learning points by a constant shift equals 0.025 in both dimensions).

Figure.1 shows the noisy data used in training the network. The data points are partitioned into 75% training set and 25% validation set. In addition, there is a test set for the final test of the performance of the models. The training set is used for the weight optimization process and the validation set is used to evaluate the network structure in the link admission process.

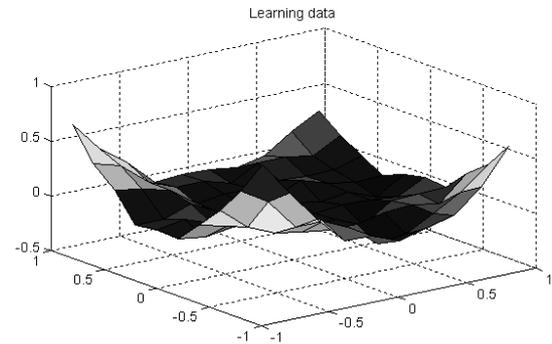

**Figure 1. Nonlinear regression problem learning data.**

The error measure used to assess the networks performance is Mean Square Error (MSE); where

$$MSE = \frac{\sum_{m=1}^{M}(y(m)-d(m))^2}{M}$$

$;y(m),d(m)$ are the network output, and the desired output at any sample "$m$" respectively; and, $M$ is the length of the investigated data set. Besides the error measures, the number of resulted networks' weights are reported to compare there complexity.

**2.3.3. Simulation Results.** Figure 2 shows the SFN network's fitting performance of a two layer SFN network. While the constructed SFN network is shown in Figure 3.

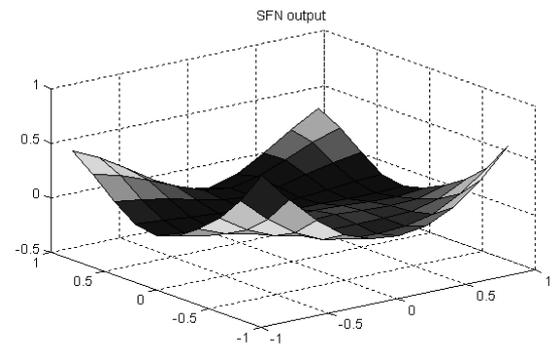

**Figure 2. Symbolic function network output.**

That is a two layer network that consists of eight elementary functions with just fourteen tuned weights. This network representation is relatively simple compared to the simplest multilayer perceptron that could be designed to model this problem that contains more than sixty weights on average [1]

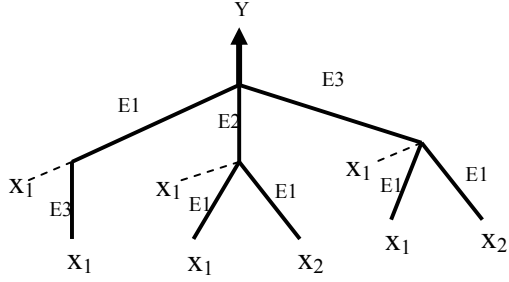

**Figure 3. The constructed symbolic function network.**

The constructed symbolic function that represents the function is:

$$Y = E1(x_1 + E3(x_1)) + E2(x_1 + E1(x_1) + E1(x_2))$$
$$+ E3(x_1 + E1(x_1) + E1(x_2))$$

$$Y = 1.878(1 + (x_1 - 0.31\log(x_1^2 + 1))^2)^{-0.592}$$
$$-0.307 e^{0.868(x_1 + 0.83(x_1^2+1)^{-0.76} + 1.89(x_2^2+1)^{-0.38})}$$
$$+ 2.607\log((x_1 - 0.316(x_1^2+1)^{-0.75} + 1.15(x_2^2+1)^{-0.72})^2 + 1)$$

The testing performance in MSE is 0.0014, while the training performance in MSE is 0.0039. This result indicates that the constructed symbolic network could capture the system nonlinearties in a very good fitness, while filtered the random noise out.

## 3. Round Trip Time prediction

### 3.1. Experimental Setup

We collected the RTT data by installing a transmission Processor at Chonbuk National University in South Korea, and a receiving-retransmission processor at Seoul National University. The transmission processor transmitted packets using the TCP-friendly mechanism. The packet size is 625 bytes, and that includes 64 bytes reserved for Probe header. The probe header keeps track of the transmission time and order of the packet. This information is used to estimate the RTT (Round Trip Time). The basic time unit is 2 Sec, so the time series of RTT values represents the measurement over each 2 Sec time interval. The RTT within each interval is the average of the individual packets' RTT. The resulted time series consists of 13,158 data points. In designing the SFN model, we used the first 2000 points as the learning data set used in designing the network, and the remaining data as an out-of- samples test data. The learning data points are partitioned into 75% training set and 25% validation set. All versions of the proposed method: 1) Forward Layer by Layer (FLY-SFN), 2) Forward Link by Link (FLK-SFN), 3) Backward (B-SFN), 4) Forward Backward (FB-SFN), and 5) Forward with Reduced Random Set (FRS-SFN) have been tested. To obtain a comparative idea about the performance of the proposed model, we have implemented a multilayer perceptron neural network (MLP) (a single hidden layer network). We have considered the following methods for training the MLP: 1) The basic backpropagation (B-BP), 2) the early stopping backpropagation (ES-BP), and 3) the Bayesian regularization backpropagation (BR-BP) [2]. It is well-known that for MLP, the most critical parameter is the number of hidden nodes. We used the validation set to determine this parameter. We trained networks with numbers of hidden nodes being 1,2,3,4,6,9,12, and 15, and selected the one that gives best validation set performance. Whether SFN or MLP, all of the network structures' performances are evaluated using a validation data set. However, only the winning structure is tested using the test data set and reported. To average out the fluctuations due to the random choice of the initial weights, we have performed 5 runs for every method, each run using different initial weights. The average testing performance is reported. For the SFN networks, all runs typically lead to the same network (and same performance). In such a case only one run's result is reported. The error measure used to assess the networks performance is Normalized Mean Square Error (NMSE %);

$$NMSE\% = \frac{\sum_{m=1}^{M}(y(m)-d(m))^2}{\sum_{m=1}^{M} d(m)^2} * 100;$$

where $y(m), d(m)$, are the network output, and the desired output at any sample "$m$" respectively; and, $M$ is the length of the investigated data set. Besides the error measures, the number of resulted networks' weights are reported to compare there complexity. In the simulation results, we used the abbreviations of the learning methods followed by the number of hidden nodes for the trained MLP network. Also, we used just the latest three data points as inputs to the network in order to predict the next data point.

## 3.2. Simulation Results

A segment of the out-of-sample data is represented in Figure 4. The Constructed SFN network consists of 6 weights and gave the performance NMSE % about 6.48 using B-SFN algorithm. The Results are listed in Table.1 that shows that the SFN could capture the RTT time series nonlinearties with a relatively simple representation than that of MLP.

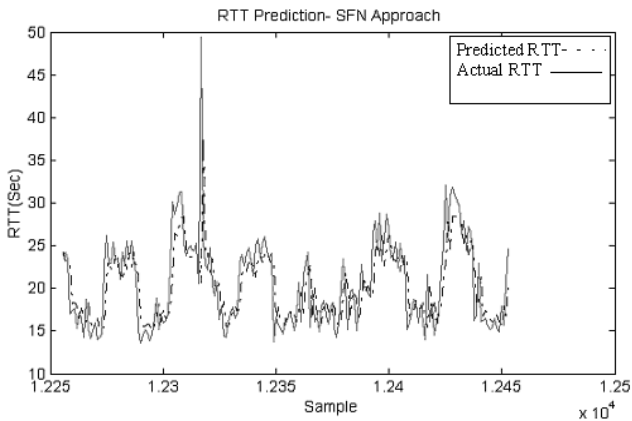

**Figure 4. Predicted and actual time series of a portion of the out-of-sample data for the SFN-RTT model.**

**Table 1. RTT predictors' testing results**

| Algorithm | Testing Average NMSE% | Average # Weights |
|---|---|---|
| B-BP (2) | 7.47 | 11 |
| ES- BP (3) | 6.53 | 16 |
| BR –BP (2) | 6.98 | 11 |
| FLY-SFN | 6.51 | 15 |
| FLK-SFN | 6.59 | 7 |
| B-SFN | 6.48 | 6 |
| FB-SFN | 6.52 | 9 |
| FRS-SFN | 6.64 | 5 |

## 4. Summary and conclusion

Prediction of the round trip time series is a hard task because the noisy and nonlinear nature of such time series. However, the SFN model could capture the RTT model dynamics and nonlinearties, and represented it in a quite simple representation that provide the fast recall speed needed by the real time predictive traffic controllers implemented in the real time transport protocols. Also, the predicted values are more accurate than the measured values because of the ability of noise filtering of the SFN model.